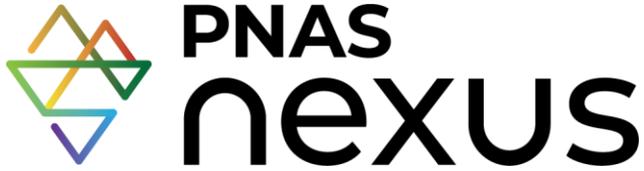

**Main Manuscript for**

# Projecting the New Body: How Body Image Evolves During Learning to Walk with a Wearable Robot

I-Chieh Lee[1,2]*, He Huang[1,2]*

1. Joint Department of Biomedical Engineering, North Carolina State University, Raleigh, NC, USA, 27695

2. Joint Department of Biomedical Engineering, University of North Carolina, Chapel Hill, NC, USA, 27599

*He Huang

**Email:** hhuang11@ncsu.edu

*PNAS Nexus* strongly encourages authors to supply an [ORCID identifier](#) for each author. Do not include ORCIDs in the manuscript file; individual authors must link their ORCID account to their *PNAS Nexus* account at https://pnasnexus.msubmit.net/cgi-bin/main.plex. For proper authentication, authors must provide their ORCID at submission and are not permitted to add ORCIDs on proofs.

**Author Contributions:** I-Chieh Lee conceived and designed the study. I-Chieh Lee performed the data analysis. He Huang acquired funding and supervised this study. All the authors wrote, reviewed and edited the manuscript.

**Competing Interest Statement:** The authors declare no conflict of interest

**Classification:** Physical Sciences and Engineering

**This PDF file includes:**

> Main Text
> Figures 1 to 6

**Abstract**

Advances in wearable robotics challenge the traditional definition of human motor systems, as wearable robots redefine body structure, movement capability, and perception of their own bodies. While these devices can empower the wearer's motor performance, there is limited understanding of how wearers update their perception of body images, especially images in dynamic movements, while learning to use these modern devices. This study aimed to fill the gap by examining the changes of body image as individuals learned to walk with a robotic prosthetic leg over multi-day training. We measured gait performance and perceived body images via Selected Coefficient of Perceived Motion (SCoMo) after each training session. Based on human motor learning theory extended to wearer-robot systems, we hypothesized that learning the perceived body image when walking with a robotic leg co-evolves with the actual gait improvement and becomes more certain and more accurate to the actual motion. Our result confirmed that motor learning improved both physical and perceived gait pattern towards normal, indicating that via practice the wearers incorporated the robotic leg into their sensorimotor systems to enable wearer-robot movement coordination. However, a persistent discrepancy between perceived and actual motion remained, likely due to the absence of direct sensation and control of the prosthesis from wearers. Additionally, the perceptual overestimation at the later training sessions might limit further motor improvement. These findings suggest that enhancing the human sense of wearable robots and frequent calibrating perception of body image are essential for effective training with lower limb wearable robots and for developing more embodied assistive technologies.

**Significance Statement**

Wearable robotics are blurring the boundaries between the human body and machines, yet how humans perceive the new body when adapting to these wearables remains unclear. This study tracked body image as individuals learned to walk with a robotic prosthetic leg. We found that wearers gradually incorporated the automated prosthesis into their internal body image, co-evolving with gait improvement. However, a persistent mismatch between perceived and actual motion, from underestimation to overestimation, was observed, potentially hindering further motor learning. These findings highlight the need for augmented sensory feedback in prostheses and perceptual calibration during physical training to enhance body image accuracy, further promoting prosthesis embodiment and acceptance and maximizing both functional and psychological outcomes for individuals with lower limb loss.

**Introduction**

Wearable robots are an emerging field designed to enhance the physical activities of users in rehabilitation settings (1, 2) or working environments (3, 4). When a robot is physically attached to a wearer to replace or augment a biological limb, it directly alters the individual's movement and their perception of their body (5-7). Existing research focuses mainly on advances in wearable robotic technologies and investigates their impact on human physical ability. Relatively limited attention has been given to address the influence of wearable robotics on human perception and cognition. In this study, we aim to investigate how a wearable robot influences a wearer's perception of body image when learning to move together with the robotic limb. The study results will contribute new knowledge in wearer-robot interactions that are important for motor learning and adaptation to wearable robotics, robotic limb embodiment, and rehabilitation and rehabilitation technology design.

**Body Image and Motor Learning in Wearable Robotics**

The ability to form a mental image of the body is a fundamental aspect of human cognition. Through sensory feedback from the physical body (i.e., proprioception, vision) (*8, 9*) and emotional states (i.e., feelings of strength or weakness) (*9, 10*), the brain integrates these factors to construct an internal body image. This body image includes not only the static size and structure of the body but also dynamic body movements in the context of interactions with the environment and objects. This cognitive ability is crucial for motor control and learning in humans (*11, 12*). Motor learning depends on the brain's ability to form internal models that predict the sensory outcomes of movements, allowing for smoother and more accurate actions (*13*). Extended from the internal model, the common coding theory suggests that perception and action share the same neural representations (*14, 15*), so effective learning requires a clear and accurate sense of one's own body, i.e., body image. A well-formed body image helps the brain anticipate what a movement should feel and look like, making it easier to adjust actions as needed. As individuals practice and gain experience, the accuracy of their body image improves, leading to better predictions, more precise movements, and faster error correction (*16*). This ongoing refinement strengthens sensorimotor integration, making body image a critical foundation for planning, executing, and adapting motor skills.

When a wearable robot is attached to the human body, it changes the wearer's physical appearance, body dynamics, and perceptions. Because wearers often do not have direct sensation and control over robots, to use these devices, wearers must learn to predict the dynamics of the device and its impact on their body. This learning process has been necessary for wearers to fully leverage the power of wearable robots to augment the motor function for successful task performance (*17, 18*). However, it is largely unknown how learning to use wearable robots reshapes the wearer's perceived body image. If the common coding theory about the perception-action coupling in motor control and learning observed in humans still holds true in wearer-robot systems, does the body image become more accurately matched to the wearer's actual motion of wearer-robot systems? Filling this knowledge gap is important for informing effective training paradigms for individuals adapting to a modern assistive device for daily activities. In addition, the new knowledge will benefit the research field of wearable robotics as body image is related to the concept of embodiment (*5, 6, 19*), which is believed to be linked to the wearer's acceptance and satisfaction with the wearable robot (*19, 20*).

**Current approach and limitation in studying body image in wearable robots**

Limited research effort has investigated body image when using wearable robotics, among which the measurement of body image is limited to the existence of body image changes or to static body image (e.g., length of the limb). Recent studies have used neuroimaging to address changes in body image within the brain after practicing the use of wearable robots (*6, 21, 22*). For example, one study (*21*) had participants learning to use an additional robotic finger to perform daily activities. After practice, the fMRI scans showed that the activation distance in the sensorimotor cortex between the biological fingers increased, indicating that the robotic thumb was integrated into the sensorimotor cortex and altered the body image. This integration influenced both motor control and the cortical finger map of the hand. Other studies have assessed body image change by asking the prosthesis wearers to indicate the changes in the size, shape, or location of their prosthetic limb(s) without visual feedback after practicing with prosthetics (*23-25*). For example, participants were asked to point to the end of their limb before and after training with a prosthetic arm or leg. (*26, 27*) Results showed a perceptual shift from the distal residual limb to the end of the prosthetic limb, indicating a restoration of body image.(*23, 25, 27*)

These existing methods for quantifying changes in body image perception with wearable robots show promise but exhibit significant limitations. These approaches are predominantly confined to static postures, relying on neuroimaging or subjective reports collected in controlled environments, rather than during dynamic movements. Consequently, most studies focus on upper-limb wearable robots, leaving lower-limb systems underexplored. Results based on static postures also raise concerns about whether findings from static conditions could translate into dynamic tasks, such as walking. Moreover, many studies rely on pre- and post-measurements, missing the benefits of multi-point assessments throughout motor learning. As a result, while existing research explains "why" wearers adapt, it provides limited insight into "what" is learned and "how" this reshaping occurs over time. A novel approach is needed to investigate how lower limb wearable robots reshape the wearers' body images during dynamic tasks like walking.

**Study Objective and Approach**

The objective of this study was to investigate how learning to walking with a robotic knee prosthesis reshapes the wearer's body image. Specifically, two main questions guided our investigation: 1) In wearer-robot systems, how does the body image of human wearers evolve over time relative to their own body motion as they learn to walk with a robotic knee prosthesis? And 2) What spatial and temporal gait parameters are likely relied upon by wearers in reconstructing their body image?

To address this goal, we conducted a four-day learning study involving non-disabled participants (n = 9) to learn walking with a robotic knee prosthesis on a treadmill (see Fig 1). The learning goal was to walk independently without touching handrails and then try to walk as fast as possible. The robotic prosthetic knee was driven by a finite-state machine that modulates joint impedance properties based on the gait phase (detailed impedance control see Appendix1). The prosthesis control parameters were manually tuned by an experienced experimenter before the practice started. More detail of the prosthesis design and control can be found in methods.

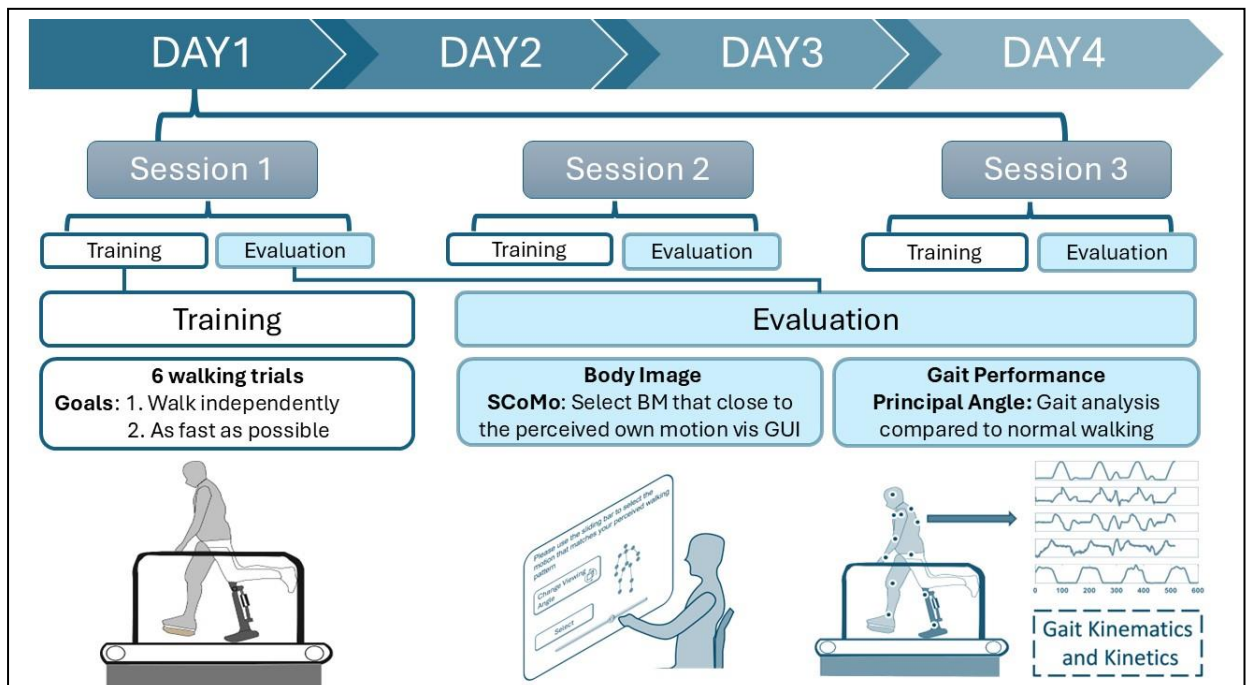

**Fig. 1. Experimental setup and protocol.** Participants were asked to learn to walk with a robotic leg on a treadmill for 4 days with a total of 12 experimental sessions. The motion of the powered knee joint was autonomously controlled using a finite-state machine impedance control strategy. On each day, three experimental sessions were performed. Each session included a training block and an evaluation block. In the training block, participants completed six two-minute walking trials. The participants were encouraged to walk without using hand railing and walk as fast as they can. In the evaluation block, the participants used a custom-designed graphical user interface (GUI) to provide SCoMo as the measurement of body image. In addition, the kinematics of participants walking with the robotic leg were captured for gait analysis to estimate the performance improvement over days.

Each training day had three practice sections (S1-S3), in which the participants walked with robotic prosthesis on a treadmill. Each session consisted of six walking trials followed by an evaluation assessment (Fig. 1). In the evaluation, we measured the participants' body image in walking by asking them to select a gait pattern displayed on the screen that best matched with their just performed walking (Fig. 2 and video 1). To do so, we introduced a new display software that can scale and synthesize various gait patterns. The Coefficient of Motion ($\alpha$) is a parameter, adjustable on the screen, that can scale the simulated gait pattern from normal walking gait to double the gait variation obtained from the participants own gait at the end of training. To measure the wearer's perceived body image, the participants were asked to adjust the coefficient α values, view the corresponding gait pattern display, and select the value that scales the gait pattern that best matches their perceived own gait pattern. The selected α value was called **S**elected **C**oefficient **o**f **Mo**tion (SCoMo) as the measurement of body image. The simulated gait pattern was displayed on the screen using biological motion (BM), composed of a set of point-light markers, each of which corresponded to the motion of a major joint of a human body (also see Fig. 2) (*28-30*). The motions of dots contained both the information of the actor's body structure (i.e., static body image) and the actor's dynamic body motion (i.e., dynamic body image). The BM display resulted in a vivid perception of walking (video 1) designed to separate biological motion information from other sources of information (such as facial expression or prosthetic limb) that are normally intermingled in human perception. This display has been used to demonstrate the invariance and distinctiveness person-specific gait fingerprint (*31, 32*), allowing viewers to recognize various attributes such as emotional status (*33*) or identity the movement refers to viewers themselves (*34*). To scale the synthesize gait patterns, we first used principal component analysis (PCA) to extract the kinematics features of the major joints' positions recorded from the participant walking with the robotic prosthesis (**P** : participant's PCA model) and a normal walking PCA model extract averaged from features of 25 non-disabled individuals (**N**: normal walking PCA model), separately (Fig. 2A). We then mathematically synthesized a BM in walking by linearly combining the walking features between the two weighted PCA models to reconstruct the 3D motion of major joints positions. The Coefficient of Motion ($\alpha$), ranged from -5 to 5, determines the ratio of the weight between the PCA models so that it can scale the synthesized different gait patterns between abnormal to normal gait. Detailed mathematical formulation can be found in the Methods section. When $\alpha = -5$, the synthesized gait pattern doubled the variation of walking with a robotic leg; when $\alpha = 0$, the gait display was the same as the participant's actual gait pattern; when $\alpha = 5$, the simulated motion represented normal walking pattern (Fig. 2). In the experiment, the α value did not show to the participants to ensure that they can only base on the displayed gait pattern of BM to make their selection.

We also measured gait biomechanics and tracked walking performance improvement along with the motor learning process over days after each practice session (Fig. 1). The gait improvement was tracked as the measurement of similarity between actual physical gait kinematics to the normal gait pattern observed in non-disabled individuals without using prostheses. The similarity of two gait patterns was quantified by the cosine of the principal angles, calculated by the subspace difference between two PCA models (P & N). This method has been used to show the similarity of high-dimensional data patterns for various applications, including muscle synergy and gait pattern (*35-37*). The smaller the value, the better was the gait pattern that was more closely resembled the normal walking. Detailed mathematics computation of gait improvement metrics based on the principal angles can be found in the Methods section.

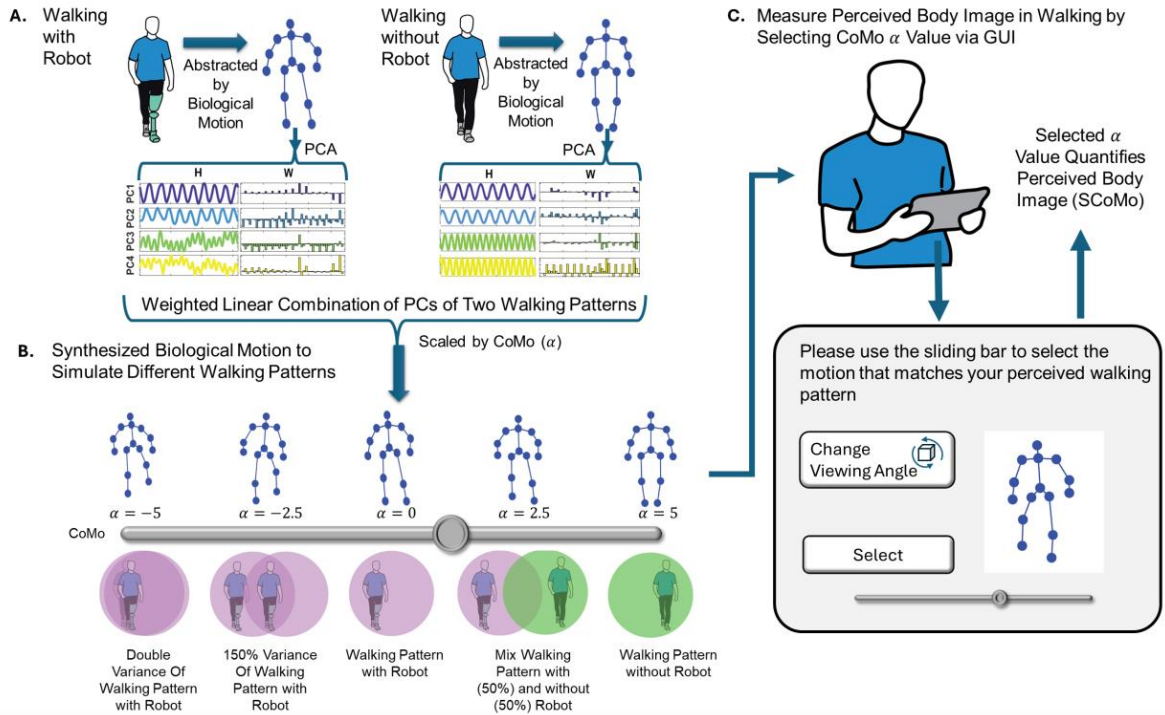

**Fig. 2. Our methodology for measuring the human wearer's perceived body image.** First, motions of participants walking with or without a robotic leg are captured and abstracted by biological motion (BM). Principal component analysis (PCA) is then applied to BM for human walking with or without wearing robotic leg, separately **(A)**. Next, the principal components that capture the basic walking patterns for either walking pattern are combined with weighted linear summation in order to synthesize a variety of walking patterns. The weights are scaled by a parameter α. Changing α∈[-5,5] leads to the synthesized BM display, ranging from the motion that doubles the variation of walking with a robotic leg to the motion that represent normal walking pattern without wearing the robotic leg **(B)**. Finally, this method in synthesizing the walking pattern is programmed into a Graphical User Interface (GUI). GUI displays the BM of walking motion while the participants choose different scaling α value via a sliding bar. To measure the perceived body image, the participants in this study were asked to select α that yields the BM that best matches their perceived self-motion. The final selected α value, called **S**elected **Co**efficient of perceived **Mo**tion (SCoMo), is the measurement of the body image.

After all data collection, a correlation analysis was performed between the body image measurement (SCoMo) and various spatial and temporal gait parameters (such as gait symmetry, trunk range of motion, etc.) This analysis provided insight into the major gait parameters that wearers may attend to when selecting their perceived body image in gait. Gait parameters with a correlation coefficient ($R^2$) greater than 0.5 were considered important cues that were associated with the wearer's perceived walking pattern and provided SCoMo value.

**Hypothesis and Significance**

We grounded our hypotheses on the assumption that the perception-action common coding theory of human motor control and learning remains applicable to wearer-robot systems. We hypothesize that the perception of body image in wearer-robot systems will co-evolve with motor learning (i.e., improvement of gait performance). Specifically, we expected that as gait performance improves during learning in walking with robotic limb, the perception of body image (including robotic limb) will become more certain (lower standard deviation of SCoMo) and more accurate in matching the actual body motion (SCoMo value was closer to zero). In addition, the correlation analysis allows us to identify the spatial and temporal gait parameters that are most strongly correlated to body image. Since processing all the details of motion is cognitively demanding (*38*), we hypothesize that certain gait parameters will be more perceptually salient, indicating that wearers selectively focus on specific motion cues when adapting to a prosthetic limb.

This project builds on growing momentum in the new field of wearable robots and addresses physical and cognitive wearer-robot interactions grounded on the theory in human motor control and learning. The study results offer new insights and knowledge on how humans perceive and integrate wearable robots into their cognition of body image during wearable robot adaptation and introduce novel methods to quantify the body image during dynamic walking. In addition, the new knowledge gained in this study can inform the better design and control of embodied wearable robots that are more acceptable by the device wearers. In addition, the study results could also guide development of more effective therapeutic interventions in the future for individuals in learning to use robotic lower limb assistive devices (such as robotic prosthetic legs and exoskeletons), which can enhance both physical mobility and mental well-being for individual wearers with motor disabilities.

## Results

### Gait Performance Improvement

Participants were asked to learn to walk with a robotic leg on a treadmill for 4 days with a total of 12 experimental sessions (Fig. 1). The goal of this study was to walk with the powered prosthetic knee as fast as possible without touching the handrails. The treadmill speed started at 0.3 m/s and was increased by 0.05 m/s after every three 2-minute walking trials until it reached 0.5 m/s, completing Day 1. Starting on Day 2, participants could request a 0.05 m/s increase, decrease or maintenance in treadmill speed if they successfully completed at least two out of three walking trials independently, without touching the handrails.

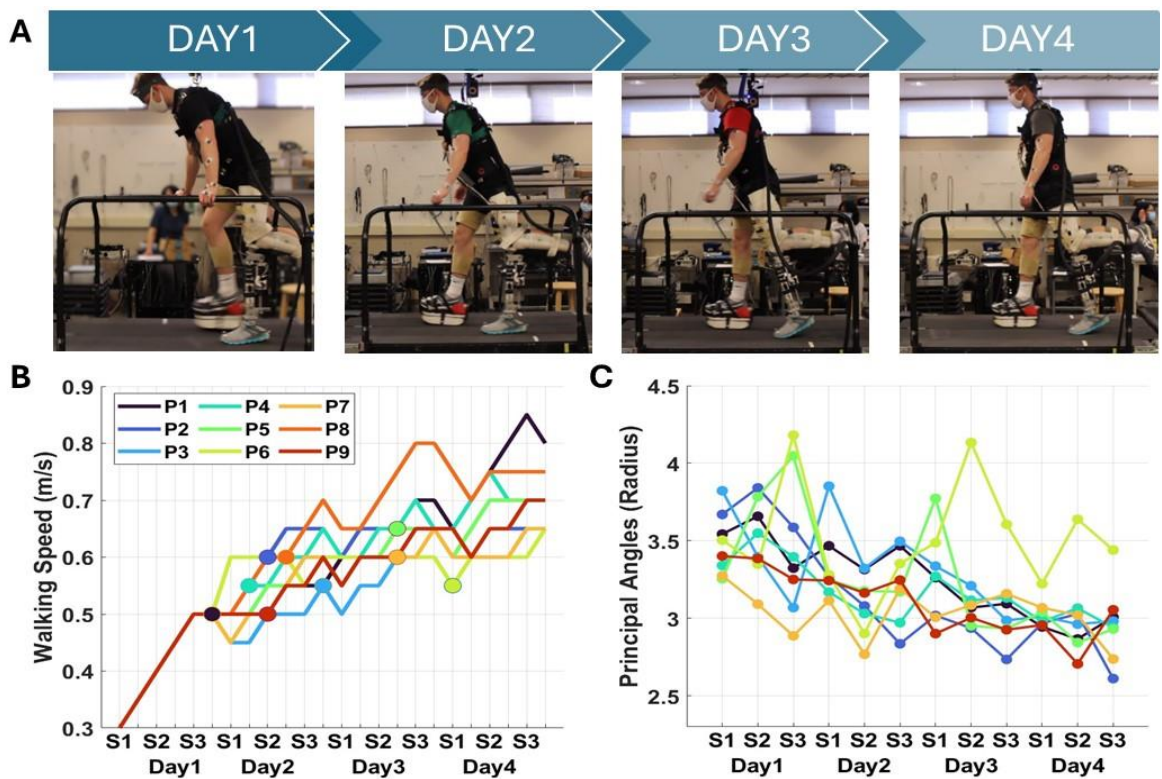

**Fig. 3. Gait improvement.** (A). Illustration of gait during the 4 days of practice from participant 2. Each photo is the screen shot of the last walking trail at each practice day. (B). Walking speed at each practice session across 4 days of practice for each participant (P1-P8). The marked circles indicated the practice session that the participants can successfully walk with the robotic leg without touching the handrails. (C). Sum of the cosine of the principal angles between the participant's gait and the normal walking PCA across all the practice sessions. The radius of 0° indicates that the participant's gait pattern is identical to the normal gait, and the larger the radius, the greater the deviation from normal walking. The decreasing trend indicates that the participants' walking patterns improved over time, approaching normal symmetrical walking. The decreasing trend indicates that the participants' walking patterns improved over time, approaching normal symmetrical walking.

To evaluate whether the gait performance improved over time, the generalized linear mixed model (GLMM) was applied. For the GLMM fitting, the participant was treated as a random effect and the practice session was regarded as a fixed effect to consider individual differences. Fig. 3A shows the improvements in walking stability and balance for one represented participant in learning to walk with the robotic prosthesis. On day 1, this participant showed poor balance and relied heavily on the handrails when walking; whereas on day 4, he walked with a much better upright posture and no longer needed the handrails. Fig. 3B shows that all participants learned how to walk with the prosthesis independently and could gradually increase their walking speed during the four days of practice. The practice sessions needed for walking without touching handrail ranged Session 3 (S3) on Day 1 for Participant 1 (P1) (the fastest learner) to Session 1 (S1) on Day 4 for P6 (the slowest learner). The fastest walking speed reached for each participant ranged from 0.85 m/s (P1 the fastest learner) to 0.6 m/s ( P6 the slowest learner). The GLMM fitting results indicated a significant increase in walking speed across practice sessions ($t = 2.047$, $p < 0.05$).

Fig. 3C illustrates the similarity of whole-body gait coordination to normal walking. Since whole-body coordination during walking involves high-dimensional data, we quantified coordination improvement by calculating the sum of the cosines of the principal angles (in radians) between the participant's and normal walking PCA models. A smaller value indicates a smaller deviation from, and thus greater similarity to, symmetrical normal walking. The decreasing trend suggests that participants' walking patterns improved over time, gradually approaching normal symmetrical walking. The GLMM results indicated a significant reduction in deviation from normal walking ($t = 2.08$, $p < 0.05$). However, we also observed that a few participants (e.g., the slower learner P6) showed fluctuations in principal angles rather than a smooth, continuous decrease.

**Measurement of Body Image During Learning**

SCoMo quantified the participants' perceived body image. Fig. 4 shows the changes in body image measurements (SCoMo) across learning sessions and days for all the participants. In the experiment, three viewing angles (Frontal View, Robotic Leg View (45° toward the robotic leg side), and CTL Leg View (45° toward the contralateral (non-robotic) leg side)) were available for participants to view the biological motion of gait, as some gait parameters might be more easily observed from one angle than another (see top panel of Fig. 4 for the viewing angles). A SCoMo value of zero represented participant's own motion recorded from the last practice trials; the positive values indicated the level of overestimation of the self-motion (i.e., the perceived body image in walking was close to normal gait than it actually was), while the negative values showed the amount of underestimation of the self-motion (i.e., the perceived body image in gait had greater variation than its actual value).

The GLMM fitting results indicated an increasing trend in the selection of SCoMo across all viewing angles (Contralateral side: t = 3.55, p<0.05; Robotic: t = 2.43, p < 0.05; Frontal: t = 3.33, p < 0.05). At the beginning of practice, the majority of participants underestimated their gait patterns, perceiving their self-motion as worse than it actually was. However, as their walking skills improved, participants tended to overestimate their own performance, perceiving their walking patterns as closer to normal walking than they actually were.

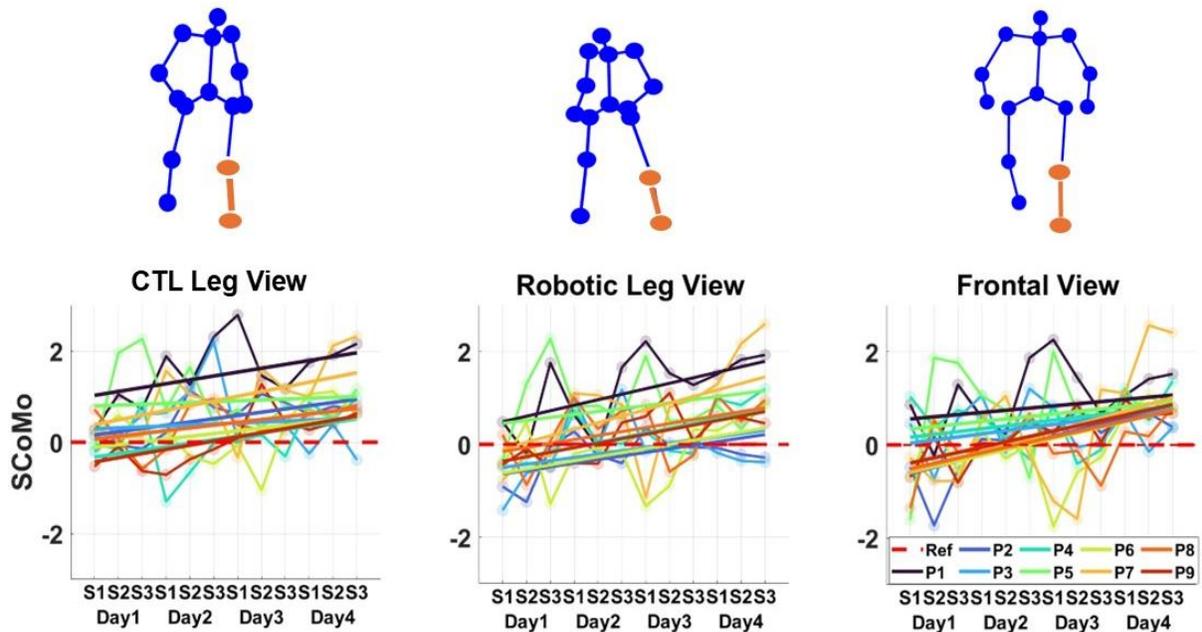

**Fig. 4. Results of SCoMo at three viewing angles over the 4 days of training.** The marked circles indicate the average SCoMo at each the practice for each participant. The fitted lines show the trend of SCoMo change over the 4 days of practice sessions for each participant. The red dashed line represents a reference (SCoMo = 0), which corresponds to the BM gait display showing the participant's own motion. Deviations from zero indicate the level of overestimation (positive values) or underestimation (negative values) of self-motion. An increasing trend in SCoMo selection suggests that participants perceived their body image as gradually improving toward normal walking.

**Consistency of perceived body image**

The body image assessment for each viewing angle was repeated six times to determine whether participants had a clear body image and could make consistent selections. The standard deviation of SCoMo was calculated to measure the perception the consistency of body image perception. Fig. 5A-5C shows the GLMM fitting of the standard deviation (SD) of SCoMo values. The significant decrease in SD over time demonstrated that participants became more consistent in their gait body image selection (Contralateral: $t = -3.9709$, $p < 0.05$; Prosthetic: $t = -6.5121$, $p < 0.05$; Frontal: $t = -6.917$, $p < 0.05$). A smaller SD implied that the selected perceived gait pattern varied less, suggesting that participants were able to consistently choose a similar gait pattern, regardless of whether they overestimated or underestimated their own motion. Furthermore, the subjective reports of confidence levels in their selections also showed an increasing trend, aligning with the SD results (see Fig. 5D).

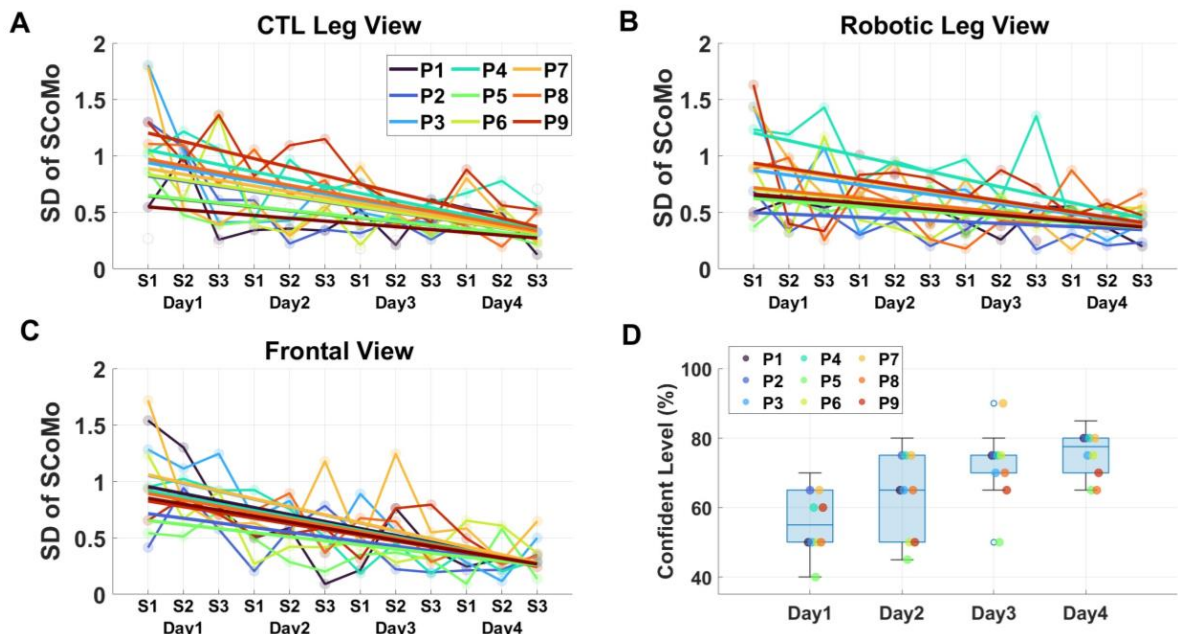

**Fig. 5. The certainty of the selection of SCoM.** Fig A-C demonstrated the result of standard deviation of SCoMo displayed at three viewing angles. The marked circles indicated the standard deviation (SD) of SCoMo from each participant. The fitted lines show the trend of SD SCoMo over the 4 days of practice sessions. The decreasing trend indicates that the participants became more certain and confident that their selection matched their body image. (D). The whisker scatter box plot shows the subjective report of confidence in selecting their own gait motion, averaged from all participants at each practice day. Participants increased their confidence in believing that the selected biological gait motion matched their own gait.

**Association between perceived body image and gait parameters**

To understand what physical gait actions were prime to perception of body image in gait, at the end of each practice day, we asked participants to name the movement parameters that they relied on to make their selection of SCoMo. Eight gait parameters were reported by the participants, including trunk medial-lateral range of motion (Trunk ML), the maximum forward leading angle between the trunk and the vertical line (Trunk Lean), robotic leg step time and step length (Robot ST and Robot SL), contralateral leg step length and step time (CTL SL and CTL ST), and the symmetry index for step time and step length (ST SI and SL SI) (Fig. 6B). We calculated these gait parameters and conducted Pearson correlation analyses between each parameter and the SCoMo values. Fig. 6B shows the coefficient of determination ($R^2$) for each gait parameter across three viewing angles, averaged across all participants. An $R^2 > 0.5$ indicates a significant correlation between the gait parameter and perceived body image (see Fig. 6A for an example).

The gait parameters strongly correlated with SCoMo selection were ST SI, Trunk Lean, CTL SL, and Robot ST. We counted the number of times each gait parameter had an $R^2 > 0.5$ across all participants and viewing angles. ST SI, CTL SL, and Robot ST exceeded this threshold in 12 out of 27 cases (9 participants × 3 viewing angles), while trunk lean did so in 10 cases. Regarding the viewing angles that provided participants with more action cues for SCoMo selection, the contralateral (CTL) view had the highest number of gait parameters used by the participants, followed by the robotic and frontal views. Specifically, the CTL view yielded 29, the robotic view 32, and the frontal view 22, out of a total of 72 counts (9 participants × 8 gait features).

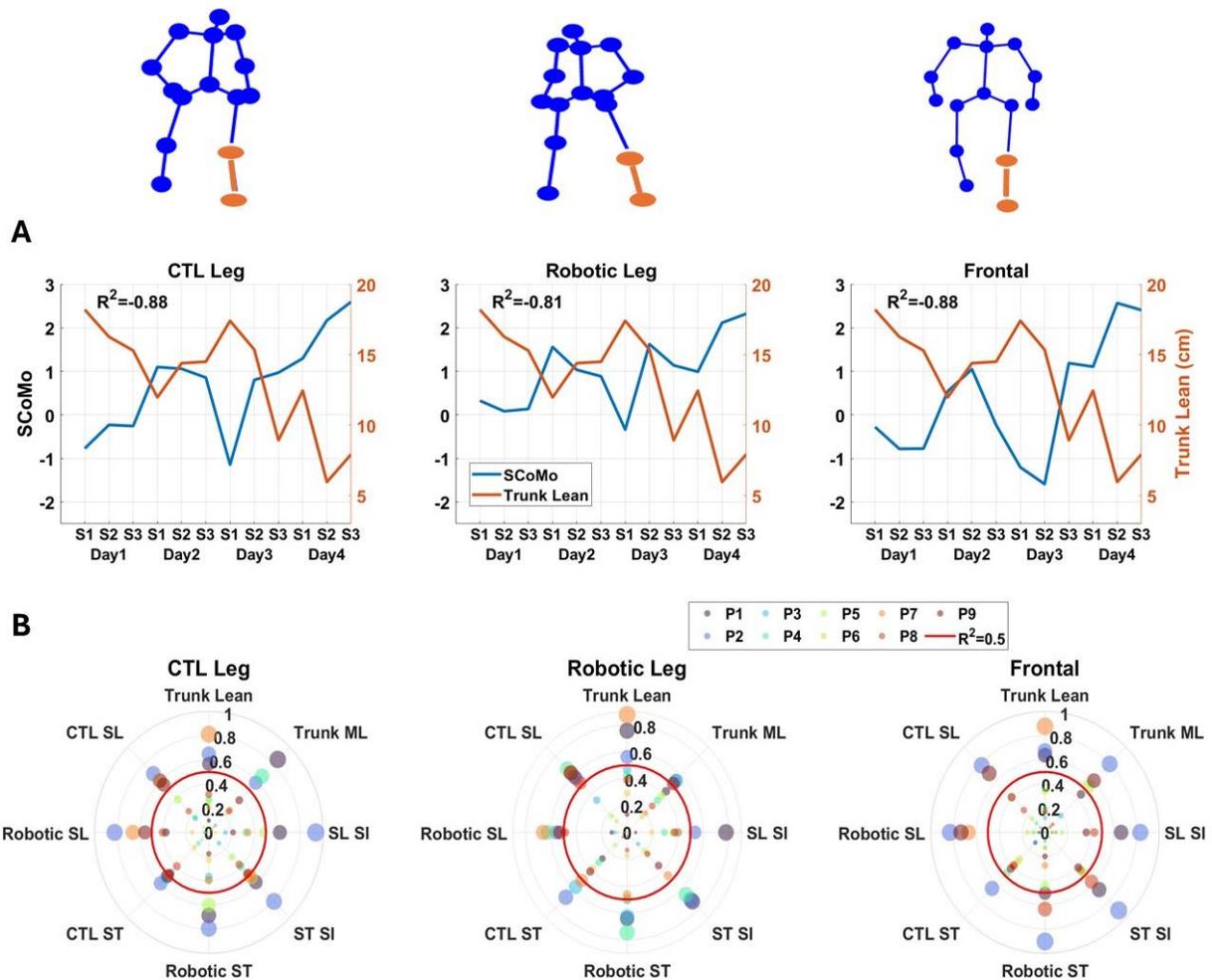

**Fig. 6. Association between SCoMo and the gait features**. (A.) An example from participant 7 (P7) showing a high negative correlation between Trunk Lean and the selection of SCoMo. When the trunk leaned less forward, the participant tended to select larger SCoMo. (B.) The polar plots show the $R^2$ between the gait parameters and the selected SCoMo at three viewing angles. The filled circles are the calculated $R^2$ from each participant with the size and location varying according to the $R^2$ value, meaning higher $R^2$ with larger circle size and away from the center of the polar plot. The red reference line indicates the $R^2 = 0.5$. The contralateral (CTL) leg viewing angle had a larger number of circles outside the reference ring. Trunk Lean, CTL Step Length (SL), Robotic Stance Time (ST), and Stance Time Symmetry Index (ST SI) are the features with a greater number of participants showing higher $R^2$ values.

## Discussion

It is believed that one's sense of presence is intimately tied up with one's physical body, and such long-lasting changes in perception can be achieved through experience-dependent motor learning (*39, 40*). Wearable robots physically attached to the wearer's body act as mediators to compensate for the body's physical limitations and potentially alter the composition of the human body with added artificial machines. Whether and how a human wearer cognitively incorporates a wearable robot into their body representation would influence their motor performance while learning to interact with wearable machines in motor tasks (5, 41). However, the related studies and evidences have been scarce. In this study, we asked non-disabled individuals to learn to walk with a robotic artificial leg and investigated (1) whether wearers incorporate a robotic limb into body image, (2) how the wearers' body image re-shape throughout the learning process, and (3) what gait features contribute to body image reformation. The outcome of this study, for the first time, directly tracks the process by which wearers regain a sense of new "body" (i.e., wearer-robot system) by quantifying the change of body image while learning to walk with the wearable robot. Through the motor learning in walking with the robotic prosthesis, all the participants improved both physical gait performance and perceived body image in gait, approaching normal, symmetrical gait. The body image became clearer and more certain to the prosthesis wearers, although the perceived body image does not necessarily match batter with the actual motion during learning.

The new evidence opens up many research directions in human motor control and human-robot interactions; it also informs the design of embodied wearable lower limb robots and novel rehabilitation interventions for training patients to efficiently use new assistive devices that can restore the mobility and mental wellbeing of the patients with physical disabilities.

**Modification of body image while learning to walk with wearable robot**

While learning to walk with a robotic leg, participants produced a qualitative change in motor performance. At the beginning of the learning, the participants had difficulty triggering or coordinating with the robotic limb's motion, failed to walk stably, and had to heavily rely on the handrails to keep balance. Through practice in several sessions, the participants clearly improved the movement coordination with robotic leg's mechanics in gait, increased walking speed, and were able to walk stably and independently on the treadmill without any balance assistance (Fig. 3A &3B). This gait pattern reorganization and performance improvement indicated that the motor learning in walking with a robotic leg leads to the integration of the robot limb as a part of the internal model in the brain to perform the action.

During learning, we also observed gradual improvement in both perceived body image (Fig. 4) and actual body motion (Fig. 3C) towards a symmetrical, normal walking pattern. This co-evolvement of both perception and action of the new "body" (i.e., wearer-robot system) in dynamic walking indicates that human wearers incorporated the robotic leg into their own body image via motor learning (*42*). Furthermore, practice in walking helped the human wearers to form a clearer body image about themselves wearing the robotic leg as the participants became more consistent and more certain in selecting a specific walking pattern as their perceived body image along the training days (Fig. 5). Having a high certainty in perceived body image ensures a clear mental reference of 'knowing where the body is, which can facilitate a reliable estimation of error between actual and desired body state to further aid in effective motion execution and learning. Overall, the updates of body image during the motor learning process are aligned with our expectation based on human motor learning theory (*43*).

**Discrepancy between perceived body image and actual physical performance**

Nevertheless, it is interesting to note that in our study, errors between perceived and actual gait pattern persisted and were not reduced during learning. This was contrary to motor learning theory which suggests that motor learning leads to a convergence between perceived body image and actual motion (*13*). The lack of error reduction between body image and actual motion indicates that having a highly precise perception of body motion may not be crucial for participants to achieve walking goals when learning to walk with an autonomous robot. Unlike aiming tasks such as grasping or reaching, walking with a robotic leg does not require adherence to a single fixed gait pattern to succeed (*44*). The observed error in this study was not sufficient to influence the task goal and was, therefore, not regulated by the wearers.

In fact, most participants initially underestimated their walking performance (SCoMo < 0) and then overestimated it (SCoMo > 0), suggesting that psychological factors may affected the body image. Emotion and attitude towards the body can be the potential factors that influences the awareness of body image (*45*). It is possible that early frustration with coordinating the robotic leg may have amplified perceptions of instability, leading to body image selections worse than the actual motion. Later in training, when participants felt more confident and accomplished, the excitement of successfully walking with the device may have biased them toward overestimating their performance. In psychology, this type of cognitive bias is known as the Dunning–Kruger effect, where individuals—especially beginners—tend to overestimate their own abilities (*46*).

Our observations may also explain the slower learning rate seen during the later stages of motor learning in this study. Motor learning is an error-based process, where individuals must recognize the difference between their estimated body state and the desired body state to optimize movement outcomes (*43, 47*). When wearers believe their movement (estimated state) are already close to normal walking (desired state), they may become less aware of, or less sensitive to, remaining performance errors, and it resulted in a plateau or saturation of gait performance. The overestimation observed in the later learning stage in this study potentially may slow down or halt further learning progress. Hence, methods to further improve body image accuracy may be needed. In this experiment, participants mainly relied on proprioception, which may be insufficient to construct accurate body image in walking (*42, 48*). A simple solution to reduce the perceptual error in motion is to provide the robot wearer addition feedback of their actual walking pattern by visual feedback (*49, 50*) or previous motion display (*51, 52*) to recalibrate their perception, which potentially can lead to more improvement in task performance after practice. This approach has the potential to improve task performance beyond what practice alone can achieve. Such feedback strategies should be further explored in future research, especially for clinical translation to support individuals with limb loss learning to walk with a new robotic prosthesis.

**Gait parameters associated with SCoMo**

When exploring the gait parameters that potentially contribute to the perception of body image, we found that the Trunk Lean, contralateral leg step length (CTL SL), robotic leg step time (Robotic ST), and step time symmetry index (ST SI) stood out with higher $R^2$ over others as they were probably the parameters that the participants were more attentive during learning (Fig. 6). These parameters are critical for completing the task goals of walking in this study, i.e., walk fast without using handrailing. The Trunk Lean angle was directly related to the use of the handrails in walking. The contralateral leg step length, which is fully controlled by the human, correlated with treadmill speed and reflected volitional motor control. In contrast, the robotic leg was autonomously controlled, and no real-time feedback of its system state was provided to participants in this study. As a result, participants likely focused more on sensing the timing of the robotic leg's motions, such as foot contact and toe-off events, to effectively coordinate their own movements with the robotic assistance.

A promising future direction would be to investigate whether providing state feedback of the robotic limb through a haptic interface (*53*) or afferent neural interface (*54*) could shift the wearer's reliance from the contralateral limb toward the robotic limb. This could potentially result in a more accurate formulation of body image and a more efficient motor learning process. We also observed that the frontal view of the biological motion display resulted in the fewest gait parameters being identified as influential for body image selection (Fig. 6), compared to the robotic and contralateral views. Since there is currently no consensus on the optimal way to visualize gait for body image assessment, we tested three viewing angles. These findings suggest that the 2D frontal display may omit critical kinematic cues available in the oblique or side views. Therefore, incorporating a variety of viewing angles in the design of future graphical user interfaces (GUIs) for body image quantification is important for capturing a more complete representation of gait.

**Implication for body image measurements and wearable robot embodiment**

Incorporation of a wearable robot into body image is considered a key component of robot embodiment (*6, 19*). However, measuring body image during dynamic movement has posed significant challenges in prior research (*6, 19*). In this study, we introduced a novel approach to measure the wearer's body image in walking by developing a continuous variable, SCoMo. This approach is very different from the existing methods that were tested and quantified in static postures and heavily relies on the questionnaire. The present body image measurement method may be extended to other wearable devices (such as exoskeletons) and other dynamic motor tasks (such as repeated reaching tasks) to investigate wearable robot embodiment.

Importantly, our research, for the first time, demonstrated that an autonomously controlled robotic prosthetic leg can also be embodied by human users. Unlike upper-limb prostheses—which often require bi-directional neural-machine interfaces to achieve embodiment—our findings suggest that, due to the automaticity and cyclic nature of walking, consistent motor practice alone is sufficient to reshape body image and internal model of the body (body schema), ultimately leading to prosthesis embodiment. This insight highlights the crucial role of clinical gait training not only in restoring motor function for individuals with limb loss, but also in facilitating the integration of the prosthesis into their body image. Body image is closely tied to prosthesis acceptance, embodiment, and mental wellbeing, making it an essential aspect of successful rehabilitation.

This involves evaluating aspects such as body image to grasp the wearer's awareness of their "new body structure" (the fusion of their body with the wearable robot) and the underlying subconscious mechanisms facilitating control and adaptation to the prosthetic device (internal body schema in preparing, planning and executing the movement).

**Future directions and Study limitations**

A natural extension of this study is how to use learned knowledge of wearer-robot interaction to advancement of robotic prosthetics technologies, which can (1) optimize and restore the body image in individuals with limb loss, thereby enhancing their confidence and mental health, and (2) reduce discrepancies between perceived and actual motion, which could facilitate more effective motor learning and improved movement performance. Several potential solutions include enhancement of wearer-prosthesis coordination through prosthesis specific therapy training (*55*) and optimizing body image (e.g., SCoMo value) by using human-in-the-loop optimization to configure the adaptive prosthesis control(*56, 57*). Finally, we are interested in translating our research knowledge to the population with lower limb loss or other potential populations who can benefit from wearable robotics for their effective use and embody of advanced wearable robotics to improve their physical and mental health and quality of life.

This study has several limitations. First, while the study results may be informative for amputee populations, we did not recruit this population in order to control the motor learning process as individuals with leg amputations might have different experiences in walking with artificial legs. For example, amputees who use a robotic prosthesis daily are already experts, leaving little room for us to study the re-shaping of body image in task learning. Second, we employed a linear Generalized Linear Mixed Model (GLMM). We acknowledge that the changes in gait performance and SCoMo did not consistently follow a linear pattern. However, due to the limited number of data points per practice session, employing a more complex model was not feasible, and a linear fit was sufficient to initially answer our research questions. Consequently, we cannot dismiss the possibility that other gait parameters may have played a crucial role in the perception of body image. We focused on the gait parameters reported by participants because it might provide a more straightforward relationship with their selection of body image.

**Conclusion**

This study addressed how humans reshape their perceived body images when wearing and adapting to wearable robotics for movement augmentation. We introduced the Selected Coefficient of Perceived Motion (SCoMo), a novel method for quantifying body image perception in walking and tracked the body image changes along with gait performance during practice. We grounded our central hypothesis on human motor learning theory, extended for wearer-robot systems, and expected that body image perception would become more certain and more accurate in matching the motor performance and its improvement across practice days. Our study reveals that the adaptation of body image stands out as a vital cognitive ability, enabling wearers to integrate the wearable robot into their motor control system with improved wearer-robot coordination in gait. However, a mismatch between perceived and actual body motion was observed, from initial underestimation to later overestimation. This might be partly due to the lack of direct and detailed sensation from the robotic prosthesis. Future work should focus on design of sensory interfacing technologies for robotic prosthetic legs and frequent recalibration of body image in physical training to further improve the function and utility of robotic prosthetic legs and to augment the motor ability and mental well-being of wearers of robotic prosthesis with limb loss.

**Materials and Methods**

*Participants*

Eight non-disabled individuals (aged: 21.2 ± 4.8 years; weight: 72.2 ± 8.0 kg; and height: 1.77 ± 0.08 m) participated in this study. The recruited participants had no known neuromuscular injuries, and no experience walking with a robotic leg that may affect their performance in this study. Participants provided written, informed consent to participate in this study approved by the University of North Carolina at Chapel Hill Institutional Review Board.

***Prosthetic Knee and Impedance Finite-State Control***

We used a robotic knee prosthesis (RKP) developed by our research group for this study. The RKP was controlled based on a finite-state impedance controller (IC) that is an established framework for robotic knee prosthesis control. The gait cycle was divided into four phases: initial double support, single support, swing flexion, and swing extension. The phase transitions were determined by knee motion and gait events (heel strike and toe-off) obtained from vertical ground reaction force, knee angle, and knee velocity(*58*). Within each phase, three impedance parameters, stiffness, equilibrium, and damping, and the real-time knee joint angle and velocity generated the knee joint torque applied to the RKP to modulate the knee motion (See Appendix1).

***Prosthesis alignment and basic movements training (Preparation Phase)***

Participants donned an L-shaped, bent-knee adapter to connect to the powered knee prosthesis. The prosthesis alignment was done following the L.A.S.A.R. protocol (*59*), and a shoe lift was used to ensure the hips were level. At the beginning of the training, we assigned pre-tuned control parameters based on the participants' body weight for them to get familiar with the RKP. Participants practiced some basic static movements (i.e., maintaining quiet standing, weight shifting), and then practiced triggering the knee motion while marching in place. The preparation phase ended when the participants could hold handrails and continuously trigger the motion 30 times. In the end, the participants practiced level-ground walking while holding handrails on a 10-meter walkway. If they could successfully trigger the prosthesis knee motion at each gait cycle, they were allowed to move to the treadmill walking experiment. The total process usually took about two separate 2 hours sessions.

*Learning treadmill walking and data collection*

After alignment and basic training, participants were asked to learn how to walk with the powered knee on a treadmill for four days. Each day consisted of 3 sessions, and each session included a training block and an evaluation block. Before each training session began, an experienced experimenter hand tuned the impedance parameters to ensure comfortable and safe walking for the participants. The training block comprised six 2-minute walking trials for participants to practice walking with the robotic leg.  On day 1, the treadmill speed started at 0.3 m/s and increased by 0.05 m/s every three trials until it reached 0.5 m/s to finish the training. After day 1, every three walking trials the participants could request to maintain or change the speed by 0.05 m/s, but an increase in speed was only allowed if two out of three trials were completed without touching the handrails. The goal was to walk as fast and stably as possible without touching the handrails. At each walking trial, a 12-camera motion capture system (Vicon, Oxford, UK) was used to record the position of reflective markers at 100 Hz and a split- belt Bertec fully instrumented treadmill (Bertec Co; Columbus, OH) was used to record the bilateral ground reaction forces at 1000Hz. We employed the Vicon full body plug-in gait model (40 reflective markers, including the sacrum) for kinematics data collection. The ground reaction force was used to determine the gait events.  Marker trajectories and ground reaction force data were synchronized, recorded, and pre-processed using Nexus 2 software. A low-pass Butterworth filter with a cut-off frequency of 6 Hz was applied to filter out the noise from the data.

The evaluation block was designed to measure body image through SCoMo. At the end of each training session, the kinematic data recorded from the participant's last walking trial and a data set from 25 non-disabled male walkers (*60*) were utilized to formulate the gait model used to synthesize the biological gait motion. The synthesized gait pattern was displayed from the frontal view, 45º rotated toward the robotic leg side and 45º rotated to the contralateral side (CTL side) to the participant. Participants were asked to freely drag the sliding bar to update the gait pattern on the screen, and then select a gait pattern that best matched with their perceived self-walking image. For each viewing angle, they repeatedly made the selection six times to measure the consistency of their selection (i.e., not a random guess). To avoid the participants following the previous selection to make decisions for all the selections, the length of the sliding bar and the initial position of the sliding bar handle were slightly varied. Addition, α value (SCoMo) did not show to the participants, so they only can be based on the displayed gait pattern of BM to make their selection. (see Fig 2 and Video 1 for the designed GUI).

*Body image test (SCoMo)*

To create the mathematical model used to synthesize the biological gait motion, we utilized two data sets, one was the gait kinematic recorded from the participant at the end of each practice session (8 continuously gait cycles), and the other one was the normal gait kinematics, data obtained from Troje (*60*) collected from 25 healthy young walkers. The basic concept was to extract the gait features from the two data sets using principal component analysis (PCA), and then proportionally mesh the extracted gait features (PCs) to reconstruct the biological gait motion.

For the participant's kinematics data, the location of markers was used to estimate 15 virtual joints center positions in 3D space (joints of the ankles, knees, hips, wrists, elbows, shoulders, and in the center of the pelvis, sternum, and head), resulting in a matrix with 45 dimensions (15 joints in 3D space = 45). Each matrix contained eight continuous gait cycles used to calculate the covariance matrix for performing the PCA to extract the eigenvalues, eigenvectors and scores. For the PCA model, the number of PCs that explained over 95% of variance of the data was used (see equation 1).

$$\mathbf{P}_{(t*45)} = \sum_{i=1}^{n} \mathbf{H}_{p(t*i)} \mathbf{W}_{p(i*45)} \quad \text{(Equation 1)}$$

where $\mathbf{P}_{t*45}$ is the reconstructed 15 joints position in 3D space (45-dimensions) with the length of 8 gait cycles (length t) from the participant's PCA. $\mathbf{H}_{p(t*i)}$ denotes the scores of the $i_{th}$ principal component with length t, and $\mathbf{W}_{p(i*45)}$ denotes the eigenvector of the $i_{th}$ principal component, and n is the number of PCs that explained over 95% of the data variances.

For the normal walking data set, the 15 joint marker trajectories in the 3D space from 25 non-disabled walkers (*60*) was used to perform PCA to obtain the $\mathbf{W}_{n(i*45)}$ and $\mathbf{H}_{n(t*i)}$ of the first 4 PCs. Given the referenced normal walking needed to be smooth and symmetrical, for the $\mathbf{H}_{n(t*i)}$, we perform the sinusoidal fit on each $H_{n(t*i)}$ obtaining the frequency and amplitude to reconstruct the $H_{s(t*i)}$. The $H_{s(t*i)}$ can be denoted in equation 2.

$$H_{s(t*i)} = c_i \sin(\omega_i t) \quad \text{(Equation 2)}$$

where t = 1,2,3….t, with a length of eight gait cycles, and $\omega_i$ is the fitted frequency of the ith PC, $c_i$ is the fitted amplitude of the ith PC. The coefficients of determination of the fits are 0.99, 0.95, 0.94, and 0.94 for the first four PCs, respectively.

Equation 3 describes the normal walking model used to reconstruct the 15 joint positions in 3D space ( $N_{(t*45)}$ ):

$$N_{(t*45)} = \sum_{i=1}^{4} Hs_{(t*i)} W_{n\ (i*45)} \quad \text{(Equation 3)}$$

To synthesize a variety of gait patterns ranged from normal walking to double the variance of the participant's own motion, we combine the two sets of PCs with a weighting term $\alpha$ and $\alpha 1$:

$$S_{(t*45)} = C_{(t*45)} + \frac{1}{5} * (5 - \alpha_1) * P_{(t*45)} + \left(\frac{1}{5} * \alpha\right) * N_{(t*45)}$$

$$\text{where } \alpha = \begin{cases} \alpha_1 & \text{if } \alpha_1 \geq 0 \\ 0 & \text{otherwise} \end{cases} \quad \text{(Equation 4 )}$$

where the reconstructed joints positions $S_{(t*45)}$ were proportionally scaled from the motion of normal walking ($\alpha_1 = 5$) to the motion with two times joint variation of the participant's motion ($\alpha_1 = -5$). The $C_{(t*45)}$ is the average of 15 joints trajectories in 3D space calculated from the participant's kinematic data set which represented the body structure of the participant.

We created a graphical user interface (GUI) to display the reconstructed $S_{(t*45)}$ (see Fig. 1 and video 1). The synthesized biological gait motion was presented on a 27- inch computer monitor. The slider changes the weighting term $\alpha$ to update the walking pattern on the monitor. The end values of the slider were randomized between 4.5 ~ 5 and -4.5 ~ -5 to randomize length of slider bar, and the initial position of the handle on the slide bar was also randomly placed.

**Data analysis**

*Computing the cosines of principal angles (Gait similarity to normal walking)*

The degree of a participant's gait pattern deviation from normal walking can be quantified based on the degree of principal angles overlap between two subspaces of PCA sets(*61*). This method has been applied to determine the similarity across muscles or movement synergies based on subspace geometry(*35-37*). The two subspaces were defined as the extracted eigenvectors sets that captured the gait pattern of the participant (column-space P of matrices) and the normal walking (column-space N of a matrices). The subspace similarity was computed using cosine of the principal angles to compare the subspaces spanned. A singular value decomposition (SVD) - based algorithm for computing cosines of principal angles can be formulated as follows.

Let columns of matrices $Q_p$ and $Q_n$ form orthonormal bases for the subspaces P and N, respectively. The reduced SVD of $Q_p^T Q_n$ is:

$$Y^T Q_p^T Q_n Z = \text{diag}(\sigma_1, \sigma_2, \sigma_3, \sigma_4), \quad 1 \geq \sigma_1 \geq \sigma_2 \geq \sigma_3 \geq \sigma_4 \geq 0 \quad \text{(Equation 5)}$$

where Y and Z both have orthonormal columns.
The principal angles can be computed as:

$$\theta_k = arccos(\sigma_k), \quad k = 1, 2, 3, 4 \quad \text{(Equation 6)}$$

while principal vectors are given by $Q_p Y_k$ and $Q_n Z_k$, and k is $k^{th}$ principal vector. The principal angles ($\theta_k$) range from 0° to 90°, where 0° denotes identical and 90° denotes orthogonal.

The distance between two subspaces was finally calculated as the sum of all the principal angles to quantify the gait deviation from the normal walking pattern based on the two PCA models used to build the SCoMo.

### *Gait* Parameters used for perception cues

To understand how perceived body image was associated with the performed gait parameters, Pearson correlation analysis was used to quantify the relationship between eight gait parameters and the participants' selection of SCoMo. We calculated the trunk medial-lateral range of motion (Trunk ML), the maximum forward leading angle between the trunk and the vertical line (Trunk Lead), robotic leg step time and step length (Robot ST and Robot SL), contralateral leg step length and step time (CTL SL and CTL ST), and the symmetry index for step time and step length (ST SI and SL SI). These eight features were subjectively reported by participants and were used to make their selections. Gait parameter(s) with a coefficient of determination ($R^2$) greater than 0.5 were considered key factors that participants relied on to form their body image.

### *Statistics*

Our first hypothesis was that the accuracy of the perceived body image would improve during learning. To answer this question, the SCoMo (α) and the practice sessions were modeled using Generalized Linear Mixed Model (GLMM) (GLMM) for each BM viewing angle (frontal, intact and prosthetic view). To determine whether gait improvement could enhance a clear body image, the standard deviation of the SCoMo was also fitted with the GLMM model to understand if participants could improve the consistency of BM selection during learning. For the GLMM fitting, the participant was treated as random effect and the practice sessions were regarded as fixed effect to take into account individual differences. The significant level was set at $p < 0.05$, and all analyses were performed using Matlab statistics toolbox.

## Data Availability Statement

The datasets generated and analyzed during the current study are available from the corresponding author upon reasonable request.

## Acknowledgments

This research was funded by the National Institutes of Health and National Science Foundation (NSF 2211739). We also thank Reilly Stafford for commenting on this paper.